%% file: camera_ready.tex
\documentclass[a4paper, 10pt, conference]{ieeeconf}      % Use this line for a4
\usepackage{FG2024}

\FGfinalcopy % *** Uncomment this line for the final submission

\IEEEoverridecommandlockouts                              % This command is only
\usepackage{graphics} % for pdf, bitmapped graphics files
\usepackage{epsfig} % for postscript graphics files
\usepackage{mathptmx} % assumes new font selection scheme installed
\usepackage{times} % assumes new font selection scheme installed
\usepackage{amsmath} % assumes amsmath package installed
\usepackage{amssymb}  % assumes amsmath package installed
\usepackage{booktabs}
\usepackage{multirow}
\usepackage{colortbl}
\usepackage{amssymb}
\usepackage{amsmath}
\usepackage{mathtools}
\usepackage{pifont}
\usepackage{hyperref}
\usepackage{url}
\newcommand{\cmark}{\ding{51}}
\newcommand{\xmark}{\ding{55}}
\usepackage{algorithm}
\usepackage{algpseudocode}
\usepackage{color}

\def\FGPaperID{228} % *** Enter the FG2024 Paper ID here

\title{\LARGE \bf
Hyp-OC: Hyperbolic One Class Classification for Face Anti-Spoofing
}

\author{\parbox{16cm}{\centering
    {\large Kartik Narayan and Vishal M. Patel}\\
    {\normalsize \{knaraya4, vpatel36\}@jhu.edu}\\
    {\normalsize
    Johns Hopkins University\\}
    {\normalsize \textcolor{magenta}{\url{https://kartik-3004.github.io/hyp-oc/}}}}
}

\begin{document}

\ifFGfinal
\thispagestyle{empty}
\pagestyle{empty}
\else
\author{Anonymous FG2024 submission\\ Paper ID \FGPaperID \\}
\pagestyle{plain}
\fi
\maketitle

%%%%%%%%%%%%%%%%%%%%%%%%%%%%%%%%%%%%%%%%%%%%%%%%%%%%%%%%%%%%%%%%%%%%%%%%%%%%%%%%
\input{sec/0_abstract}
\input{sec/1_intro}

\input{sec/2_related_work}

\input{sec/3_proposed}
\input{sec/4_experiments}

\input{sec/5_results}

\input{sec/6_conclusion}

%%%%%%%%%%%%%%%%%%%%%%%%%%%%%%%%%%%%%%%%%%%%%%%%%%%%%%%%%%%%%%%%%%%%%%%%%%%%%%%%

{\small
\bibliographystyle{ieee}
\bibliography{egbib}
}

\end{document}

%% file: sec/0_abstract.tex
\begin{abstract}
Face recognition technology has become an integral part of modern security systems and user authentication processes. However, these systems are vulnerable to spoofing attacks and can easily be circumvented. Most prior research in face anti-spoofing (FAS) approaches it as a two-class classification task where models are trained on real samples and known spoof attacks and tested for detection performance on unknown spoof attacks. However, in practice, FAS should be treated as a one-class classification task where, while training, one cannot assume any knowledge regarding the spoof samples a priori. 
In this paper, we reformulate the face anti-spoofing task from a one-class perspective and propose a novel hyperbolic one-class classification framework. To train our network, we use a pseudo-negative class sampled from the Gaussian distribution with a weighted running mean and propose two novel loss functions: (1) Hyp-PC: Hyperbolic Pairwise Confusion loss, and (2) Hyp-CE: Hyperbolic Cross Entropy loss, which operate in the hyperbolic space. Additionally, we employ Euclidean feature clipping and gradient clipping to stabilize the training in the hyperbolic space.   
To the best of our knowledge, this is the first work extending hyperbolic embeddings for face anti-spoofing in a one-class manner. With extensive experiments on five benchmark datasets: Rose-Youtu, MSU-MFSD, CASIA-MFSD, Idiap Replay-Attack, and OULU-NPU, we demonstrate that our method significantly outperforms the state-of-the-art, achieving better spoof detection performance.
\end{abstract}

%% file: sec/1_intro.tex
\section{Introduction}
\label{sec:intro}
\input{figures/visual_abstract}

%-------------------------------------------------------------------
We are in an era where facial recognition is extensively utilized for authentication and access control. It is used in diverse sectors, including mobile device security, financial services, border control, fraud prevention, e-commerce, healthcare, etc. However, such widespread adoption of facial recognition technology has made it vulnerable to spoofing attacks. Malicious actors attempt to deceive the system by spoofing the identity of an individual using presentation attack instruments (PAI). They employ various attacks, such as printed photos, replayed videos, or 3D synthetic masks, that jeopardize security and endanger face as a biometric modality. Hence, it is crucial to develop robust face anti-spoofing (FAS) techniques that can counter this threat.

%-------------------------------------------------------------------
\textbf{Why Unimodal FAS ?}
With the advent of sophisticated hardware, there was an influx of multimodal FAS techniques~\cite{zhang2012face, parkin2019recognizing, shen2019facebagnet, liu2021data, kuang2019multi, liu2021face, deng2023attention} that incorporate auxiliary data like depth map~\cite{yu2021revisiting, liu2018learning}, reflection map~\cite{zhang2021structure}, infrared images~\cite{zhang2019feathernets}, r-PPG signals~\cite{Hernandez-Ortega_2018_CVPR_Workshops, liu20163d}, and additional sensors~\cite{sepas2018light} to boost the performance. However, relying on such advanced hardware and sensors is problematic as they're expensive and not universally available where FAS systems are deployed. In this work, we focus on unimodal FAS, which uses the widespread RGB camera found in nearly all mobile devices and is easily accessible. It's not only affordable but also straightforward to integrate at various security checkpoints.

%-------------------------------------------------------------------
\textbf{Why One Class ?}
The FAS problem has been approached in different ways in the literature. Binary classifiers~\cite{atoum2017face, kim2019basn, wang2022patchnet, yu2020searching, george2019deep} associate samples with real and spoof labels. Domain Adaptation methods~\cite{li2018unsupervised, wang2019improving, wang2020unsupervised, quan2021progressive, zhou2019face, zhou2022generative} utilize target domain data to bridge the gap between source and target domains. Domain Generalization techniques~\cite{shao2019multi, wang2020cross, jia2020single, liu2021adaptive, srivatsan2023flip, sun2023rethinking} focus on minimizing the distribution discrepancies between multiple source domains that generalize better to unseen domains. However, all these formulations assume some kind of prior knowledge of spoof samples while training. In the real world face anti-spoofing scenario, spoof samples are infinitely variable, which makes the task of FAS inherently complex. The variations in spoof attacks are boundless. Whether you consider webcam, masks, print, surveillance camera, or phone-based attacks, malicious actors exploit variations in factors like - light, camera sensor, printer, paper, and mask to fool the facial recognition system. The vast spectrum of possibilities highlights the need to address face anti-spoofing as an anomaly detection or one-class classification problem where one only have access to the real samples, as it focuses on identifying genuine samples while remaining resilient to the ever-expanding range of spoofing techniques. Some recent works~\cite{arashloo2017anomaly, fatemifar2019combining, nikisins2018effectiveness, baweja2020anomaly}, in agreement with our approach, have formulated FAS as a one-class classification task, demonstrating its complexity and practical relevance for real-world applications.

%-------------------------------------------------------------------
\textbf{Why Hyperbolic ?}
Recently, hyperbolic embeddings are adopted for vision tasks such as image segmentation~\cite{atigh2022hyperbolic}, instance segmentation~\cite{weng2021unsupervised}, few-shot classification~\cite{gao2021curvature} and image retrieval~\cite{ermolov2022hyperbolic}.  ~\cite{nickel2017poincare, khrulkov2020hyperbolic} established that hyperbolic embeddings can outperform the Euclidean embeddings significantly on data with latent hierarchies, both in terms of representation capacity and generalization ability. In the FAS scenario, the real and spoof classes have subtle visual differences and lie close to each other in the feature space. Therefore, it is challenging to fit a hyperplane for one-class classification, especially in the absence of a spoof class while training. Hyperbolic space with a negative curvature allows for the learning of discriminative features owing to the nature of exponential growth in volume with respect to its radius. Consequently, hyperbolic space aids in learning a separating \textit{gyroplane} (Section~\ref{sec:hyperbolic}) for effective one-class FAS (See Figure~\ref{fig:visual_abstract}). 

%-------------------------------------------------------------------
In this work, we extend the use of hyperbolic embeddings for face anti-spoofing. We formulate the FAS problem as a one-class classification due to its practicality for real-world deployment and propose two novel loss functions to train our network. The following are the main contributions of our research:
\begin{itemize}
    \item We propose using hyperbolic embeddings for one-class face anti-spoofing. We show that using hyperbolic space helps learn a better decision boundary than the Euclidean counterpart, boosting the FAS performance.  
    \item We propose a novel Hyperbolic Pairwise Confusion Loss (Hyp-PC) that operates in the hyperbolic space. It induces confusion within the hyperbolic feature space, effectively stripping away identity information. Such disruption of features helps to learn better feature representations for the FAS task.
    \item We propose a novel Hyperbolic Cross Entropy loss (Hyp-CE). It uses hyperbolic softmax logits and penalizes the network for every misclassification. 
\end{itemize}

%% file: figures/visual_abstract.tex
\begin{figure}[t]
  \centering
   \includegraphics[width=1\linewidth]{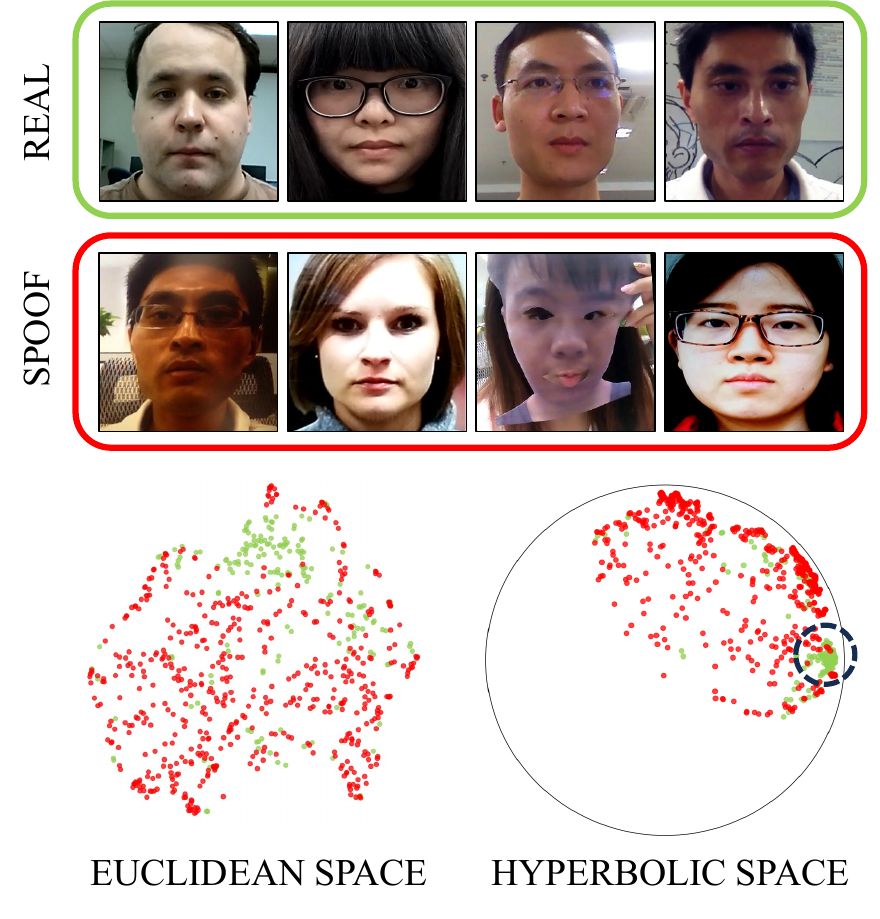}
    \caption{Feature representation of real and spoof samples in the Euclidean and the hyperbolic space. The representation of real samples in the hyperbolic space is compact (dotted circle), resulting in a better separating \textit{gyroplane} contrary to the Euclidean space in which the representation is scattered. Hyperbolic embeddings prove to be effective in one-class classification for face anti-spoofing.}
   \label{fig:visual_abstract}
\end{figure}

%% file: sec/2_related_work.tex
\section{Related Work}
\label{sec:related_work}
In this section, we give an overview of previous works in FAS, with a focus on one-class FAS. 
In addition, we briefly review the literature on hyperbolic embeddings.

%-------------------------------------------------------------------
\subsection{Face Anti-spoofing}
Earlier works~\cite{li2016generalized, de2013lbp, komulainen2013context, patel2016secure} utilized handcrafted features for FAS. Some works are based on eye blinking~\cite{pan2007eyeblink, jee2006liveness}, head movements~\cite{wang2009face, bao2009liveness}, gaze tracking~\cite{ali2012liveness}, and remote physiological signals~\cite{yu2019remote}. Classical handcrafted features such as LBP~\cite{boulkenafet2015face}, SIFT~\cite{patel2016secure}, and HOG~\cite{komulainen2013context} were leveraged to extract spoofing patterns. Subsequently, deep learning-based models~\cite{asim2017cnn, yang2019face, atoum2017face} are utilized to detect spoofs. The majority of these methods approach it as a binary classification problem. However, in practice, FAS should be considered a one-class classification (OCC) task.~\cite{arashloo2017anomaly} showed that two-class methods can be biased towards spoof samples in the training set. Following that, several works were proposed for one-class face anti-spoofing (OC-FAS) even though the performance was not competitive compared to binary FAS. An observation is the utilization of classical one-class classifiers such as OC-SVM, OC-GMM, and MD for final classification.~\cite{nikisins2018effectiveness} uses IQM features with one-class GMM for detecting spoofs.~\cite{fatemifar2019spoofing} shows identity information can be used to improve OC-FAS performance.~\cite{fatemifar2019combining} uses an ensemble of one-class classifiers.~\cite{perez2019deep} uses metric learning, where a triplet focal loss is used as a regularizer.~\cite{liu2019deep} proposed a Deep Tree Network (DTN) for zero-shot FAS.~\cite{george2020learning} uses center loss for a compact representation of the bonafide class while being away from the embeddings of the attacked class.~\cite{baweja2020anomaly} samples pseudo-negative samples from Gaussian to train the OC-FAS classifier.~\cite{george2020learning} introduces a multi-channel neural network for learning one-class representations. Despite these efforts, the performance of OC-FAS methods still lags significantly behind that of binary FAS approaches.

%-------------------------------------------------------------------
\subsection{Hyperbolic Embeddings}
%hyperbolic embeddings
Hyperbolic spaces have gained much attention for their representation capability in a wide range of domains~\cite{law2019lorentzian, sarkar2011low, yu2019numerically}. It benefits vision applications~\cite{chami2019hyperbolic, liu2020hyperbolic, long2020searching} because natural images often exhibit hierarchical structure~\cite{khrulkov2020hyperbolic, mathieu2019continuous}. ~\cite{ganea2018hyperbolic} proposed several models in the hyperbolic space, such as Hyperbolic Neural Networks, Multinomial Logistic Regression, Fully-Connected and Recurrent Neural Network.~\cite{shimizu2020hyperbolic} introduces hyperbolic convolutional layers.~\cite{guo2022clipped} performs Euclidean feature clipping to solve the vanishing gradient problem of hyperbolic networks.
Following previous works~\cite{nickel2017poincare, khrulkov2020hyperbolic}, we employ the Poincar\'e Ball model in which our proposed loss functions operate.

%% file: sec/3_proposed.tex
\section{Proposed Work}
We propose two novel loss functions: (1) Hyp-PC: Hyperbolic Pairwise Confusion loss, and (2) Hyp-CE: Hyperbolic Cross Entropy loss, both of which take advantage of hyperbolic space's capability to efficiently represent data. We employ a hyperbolic classifier head (Hyp-OC) that performs hyperbolic softmax-regression and use the resulting hyperbolic logits for one-class face anti-spoofing. An overview of the proposed framework is depicted in Figure~\ref{fig:framework}. The following section is structured as follows. Initially, we lay out the required foundational concepts of hyperbolic spaces in Section~\ref{sec:hyperbolic}. We then explain Hyp-PC loss in Section~\ref{sec:Hyp-PC} and Hyp-CE loss in Section~\ref{sec:Hyp-CE}. Lastly, we provide a comprehensive overview of the training framework and the strategies we follow to stabilize hyperbolic training in Section~\ref{sec:training}.

\subsection{Preliminaries: Hyperbolic Embeddings}
\label{sec:hyperbolic}
An $d$-dimensional hyperbolic space $\mathbb{H}^d$ is a smooth Riemannian manifold with a constant negative curvature $-c (c>0)$. There are several isometric models in the hyperbolic space, however, we operate in the Poincar\'e model~\cite{nickel2017poincare} due to its widespread usage in computer vision. The Poincar\'e ball model ($\mathbb{B}_{c}^d, g^{\mathbb{B}_{c}})$ with manifold $\mathbb{B}_{c}^{d} = \{x \in \mathbb{R}^d : c\lVert x \rVert < 1, c \geq 0 \}$ depicted in Figure~\ref{fig:poincare} is an $d$-dimensional ball equipped with Riemannian metric: 
\setlength{\belowdisplayskip}{0pt} \setlength{\belowdisplayshortskip}{0pt}
\setlength{\abovedisplayskip}{0pt} \setlength{\abovedisplayshortskip}{0pt}
\begin{equation}\label{eqn:metric}
        g_{x}^{\mathbb{B}_{c}} = (\lambda_{x}^{c})^2 g^{E} = \frac{2}{1 - c||x||^2} \mathbb{I}^{d},
\end{equation}
where $\lambda_{x}^{c} = \frac{2}{1 - c\lVert x \rVert^2}$ is the \textit{conformal factor}, $g^E = \mathbb{I}_d$ is the Euclidean metric tensor and $c$ is the curvature of the hyperbolic space. In the Euclidean space, the volume of an object with diameter $r$ increases polynomially, however, in the hyperbolic space, these volumes grow at an exponential rate because of $\lambda_{x}^{c}$ which approaches infinity near the boundary of the ball. This property allows efficient embedding of data in low dimensions.  

Euclidean vector operations are not valid in hyperbolic spaces, and operations from gyrovector spaces are adopted to operate in hyperbolic spaces. Some of the basic operations in hyperbolic spaces using the gyrovector formalism~\cite{ungar2005analytic, ungar2022gyrovector} are: 

\noindent\textbf{M\"obius Addition.}
Vector addition of two points $u, v \in \mathbb{B}_{c}^d$ is formulated using M\"obius addition as,
\begin{equation}\label{eqn:mobius_add}
        u \oplus_c v = \frac{(1 + 2c\langle u, v \rangle + c\|v\|^2)u + (1 - c\|u\|^2)v}{1 + 2c\langle u, v \rangle + c^2\|u\|^2\|v\|^2}
\end{equation}
where, $\langle \cdot \rangle$ denotes the Euclidean inner product. $\lim_{c \to 0}\oplus_c$ converges to standard $+$ in the Euclidean space.

\noindent\textbf{Exponential map.}
The exponential map $\exp_x^c$ projects vectors from Euclidean space into the Poincar\`e Ball. Euclidean space corresponds to the tangent space $\mathcal{T}_{x}\mathbb{B}_c^d$ of the manifold $\mathbb{B}_c^d$ at a reference point $x$. In our work, we treat the starting point $x$ in the Poincar\'e Ball as a parameter and optimize it using Riemannian gradient~\cite{nickel2017poincare}. For any point $x \in \mathbb{B}_c^d$, the exponential map $\exp_x^c : \mathcal{T}_x\mathbb{B}^d_c \to \mathbb{B}^d_c$ for $u$ is defined as,
\begin{equation}\label{eqn:exponential_map}
    \exp_x^c(u) = x \oplus_c \left( \tanh\left( \frac{\sqrt{c} \|u\|}{2} \right) \frac{u}{\sqrt{c}\|u\|} \right)
\end{equation}
As $\lim_{c \to 0}\exp_x^c(u) = x + u$, i.e. the exponential map converges to standard translation operation in Euclidean space.

\noindent\textbf{Distance Measure.}
The distance between two vectors $u, v \in \mathbb{B}_c^d$ in the Poincar\'e Ball is the length of the \textit{geodesic} connecting the two vectors, which is the shortest curve between those points in $(\mathbb{B}_c^d, g^{\mathbb{B}_c})$ and is defined as:
\begin{equation}\label{eqn:distance_measure}
    D_{\text{hyp}}(u,v) = \frac{2}{\sqrt{c}} \text{arctanh}\left(\sqrt{c}\|- u \oplus_c v\|\right)
\end{equation}
When $c \to 0$, geodesics becomes straight-lines recovering Euclidean geometry: $\lim_{c \to 0}D_{\text{hyp}}(u,v) = 2\|u-v\|$.

\input{figures/poincare}
\input{figures/framework}
\input{sec/algorithm}

\noindent\textbf{Hyperbolic Softmax.}
For $p \in \mathbb{B}_c^d$, $a \in \mathcal{T}_p\mathbb{B}_c^d \backslash \{0\}$, Ganea \textit{et al.}~\cite{ganea2018hyperbolic} describes the \textit{gyroplane}, i.e. the hyperplane in the Poincar\'e Ball, as:
\begin{equation}\label{eqn:gyroplane}
    \Tilde{H}_{a,p}^c := \{x \in \mathbb{B}_c^d \colon \langle -p \oplus_c x, a\rangle = 0 \},
\end{equation}
where $x$ is a hyperbolic feature vector mapped using Equation~\ref{eqn:exponential_map}. $\Tilde{H}_{a,p}^c$ can be interpreted as the union of images of all geodesics in $\mathbb{B}_c^d$ orthogonal to $a$ and containing $p$. Given $K$ classes and $k \in \{1,...,K\}$, $p_k \in \mathbb{B}_c^d$, $a_k \in \mathcal{T}_{p_k}\mathbb{B}_c^d \backslash \{0\}$, the hyperbolic distance of $x$ to the gyroplane of class $k$ is given as:
\begin{equation}\label{eqn:gyroplane_distance}
\nonumber     d_c(x, \Tilde{H}_{a_k,p_k}^c) = \frac{1}{\sqrt{c}} \sinh^{-1} \left( \frac{2\sqrt{c}\left| \left\langle -p_k \oplus_c x, a_k \right\rangle \right|}{(1 - c\left\| -p_k \oplus_c x \right\|^2)\left\| a_k \right\|} \right).
\end{equation}
The logit of class $k$ for hyperbolic feature vector $x$ is based on the distance defined in Equation~\ref{eqn:gyroplane_distance} and calculated using the Reimannian metric described in Equation~\ref{eqn:metric} as:
\begin{equation}\label{eqn:logits}
     \zeta_{p_k} (x) = \lambda_{p_k}^c \|a_k\| d_c(x, \Tilde{H}_{a_k,p_k}^c).
\end{equation}
As a result, the likelihood is given as:
\begin{equation}\label{eqn:likelihood}
      p(y = k \mid x) \propto \text{exp}(\zeta_{p_k} (x)).
\end{equation}
In our work, we utilize the above likelihood (i.e. Equation~\ref{eqn:likelihood}) as target class classification probability to calculate the performance metrics. Additionally, we use the hyperbolic-softmax logits (Equation~\ref{eqn:logits}) to calculate the Hyp-CE loss described in Section~\ref{sec:Hyp-CE}. The trainable parameters of hyperbolic classifier head are the vectors $\{ p_k\}$ and $\{ a_k\}$ for each class $k$. We refer to this hyperbolic classifier head as Hyp-OC.

\subsection{Hyp-PC: Hyperbolic Pairwise Confusion loss}
\label{sec:Hyp-PC}
In our proposed method, we perform one class training, in which each iteration consists of $n$ positive samples $\ni 2 \mid n$. Let $\mathbb{S}_n = \{S_{n}^{i} \mid 0\leq i<n\}$ be the set of hyperbolic feature vectors of the positive samples. Using the \textit{geodesic} distance formulation given in Equation~\ref{eqn:distance_measure}, we introduce a novel loss function: Hyp-PC which is defined as:
\begin{equation}\label{eqn:hyppc}
    \mathcal{L}_{\text{Hyp-PC}} = \sum\limits_{i=0}^{n/2 - 1} \frac{2}{\sqrt{c}} \text{arctanh}\bigl( \sqrt{c} \| -S_n^i \oplus_c S_n^{i + n/2} \| \bigr) \Big/ n.
\end{equation}
Here, $\| \: \|$ represents the Euclidean norm, $\oplus_c$ is calculated using Equation~\ref{eqn:mobius_add}, and ${S_{n}}^{i}$ are hyperbolic feature vectors exponentially mapped from $\mathcal{T}_{x}\mathbb{B}_c^d$ to $\mathbb{B}_c^d$ using Equation~\ref{eqn:exponential_map}. The Hyp-PC loss is utilized to refine the features of positive samples by removing identity-related information. This process declusters the positive class feature representation, enhancing the model's ability to generalize. Such an approach is particularly beneficial for the FAS task, where the focus is solely on spoof detection rather than recognizing the individual identities.

\subsection{Hyp-CE: Hyperbolic Cross Entropy loss}
\label{sec:Hyp-CE}
We propose a novel Hyp-CE -- Hyperbolic Cross Entropy loss based on logits from the hyperbolic space calculated using Equation~\ref{eqn:logits}. Given $K$ classes and $k \in \{1,...,K\}$, $p_k \in \mathbb{B}_c^d$, $a_k \in \mathcal{T}_{p_k}\mathbb{B}_c^d \backslash \{0\}$, let the mini-batch size be $2n$, the set of hyperbolic features obtained after the  \textit{exponential map} (Equation~\ref{eqn:exponential_map}) be $\mathbb{S}_{2n} = \{S_{2n}^{i} \mid 0\leq i<2n\}$, the target labels for the mini-batch be $\mathbb{Y}_{2n} = \{Y_{2n}^{i} \mid 0\leq i<2n, \dim(Y_{2n}^i) = K\}$ and the weightage of each sample in the mini-batch be $\mathbb{W}_{2n} = \{W_{2n}^{i} \mid 0\leq i<2n\}$, the Hyp-CE loss for the minibatch is defined as:
\begin{equation}\label{eqn:hypce}
    \mathcal{L}_{\text{Hyp-CE}} = \sum\limits_{i=0}^{2n - 1} \sum\limits_{k=1}^{K} -W_{2n}^i \log \frac{\exp(\zeta_{p_k}(S_{2n}^i))}{\sum\limits_{k=1}^{K} \exp(\zeta_{p_k}(S_{2n}^i))} \cdot Y_{2n}^i.
\end{equation}
In our work, we have two classes and give equal weightage to all samples. The predicted probability of $S_{2n}^i$ belonging to class $k$ ($k=0$ for real sample and $k=1$ for spoof sample) can be expressed as:
\begin{equation}\label{eqn:hypce_logits}
    z_{k=0,1} = \frac{g_k}{g_0 + g_1} = \frac{\exp(\zeta_{p_k}(S_{2n}^i))}{\exp(\zeta_{p_0}(S_{2n}^i)) + \exp(\zeta_{p_1}(S_{2n}^i))}.
\end{equation}
Using Equation~\ref{eqn:hypce} and Equation~\ref{eqn:hypce_logits} the Hyp-CE loss function can simply be written as:
\begin{equation}\label{eqn:hypce_binary}
    \mathcal{L}_{\text{Hyp-CE}} = \sum\limits_{i=0}^{2n - 1} y_i \cdot \log(z_i) + (1 - y_i)(1 - z_i),
\end{equation}
where, $y_i$ is the ground truth label and $z_i$ is the predicted probability for the spoof class.

\subsection{Training framework for one-class classification}
\label{sec:training}
Our proposed pipeline uses only real samples for training and can be viewed as a conjunction of an Euclidean network: $\text{E}(x)$, and a hyperbolic classifier head: $\text{H}(x)$. The Euclidean network comprises of two parts: (1) $\text{E}_1(x)$ - which is a facial feature extractor and (2) $\text{E}_2(x)$ which is a fully-connected neural network used for dimensionality reduction. The hyperbolic classifier head outputs the final target class classification probability explained comprehensively in Section~\ref{sec:hyperbolic}. In each batch, given a set of $n$ training images $\mathbb{X}_n = \{X_n^i \mid 0 \leq i < n\}$, we employ $\text{E}_1(x)$ to produce a set of $d^\prime$-dimensional feature set $\mathbb{F}_n = \{F_n^i \mid 0 \leq i < n, \dim(F_n^i) = d^\prime\}$. Inspired from the work of Baweja \textit{et al.}~\cite{baweja2020anomaly}, we define a Gaussian distribution $\mathcal{N}(\mu, \sigma I)$ with a $d^\prime$-dimensional adaptive mean to sample pseudo-negative features in proximity of the real samples' features. In each iteration, we sample $n$ pseudo-negative features from the distribution with mean $\mu = \alpha\mu_{\text{previous}} + (1 - \alpha)\mu_{\text{current}}$, where $\mu_{\text{previous}}$ is the mean of real samples' features of the previous batch, $\mu_{\text{current}}$ is the mean of real samples' features of the current batch and $\alpha$ is a hyper-parameter. We concatenate the real samples' features and pseudo-negative features and feed it to $\text{E}_2(x)$ for dimensionality reduction. The concatenated batch of size $2n$ when given as input to $\text{E}_2(x)$ outputs a low-dimensional representation of features $\mathbb{G}_{2n} = \{G_{2n}^i \mid 0 \leq i < 2n, \dim(G_{2n}^i) = d \}$. $\mathbb{G}_{2n}$ is then mapped to the Poincar\'e Ball $(\mathbb{B}_{c}^{d}, g^{\mathbb{B}_{c}})$ with manifold $\mathbb{B}_{c}^{d} = \{x \in \mathbb{R}^{d} : c\lVert x \rVert < 1, c \geq 0\}$ using \textit{exponential map}. Let the vectors in the hyperbolic space be represented as a set $\mathbb{S}_{2n} := \{S_{2n}^{i} \mid 0\leq i<2n\}$. Finally, $\mathbb{S}_{2n}$ is given as input to $\text{H}(x)$ that returns $\zeta_{p_k} (S_{2n}^{i})$, for $k=0,1$. These hyperbolic-softmax logits are then used for final prediction. 

We use the VGG-16 pretrained on VGGFace~\cite{parkhi2015deep} as $\text{E}_1(x)$ and a neural network with four FC layers as $\text{E}_2(x)$. The last three convolution layers and the last FC layer of $\text{E}_1(x)$ are updated during the training phase to make the feature representations suitable for the FAS task. Additionally, $\text{E}_2(x)$, the starting point $x$ on the Poincar\'e Ball (described in Section~\ref{sec:hyperbolic}) and $\{ a_k\}$, $\{ p_k\}$ vectors of the hyperbolic classifier head are trainable parameters of the proposed pipeline which are updated when the loss is backpropagated. The proposed loss functions operate using the hyperbolic features. The Hyp-PC loss uses  $\mathbb{S}_{n}$ and Hyp-CE loss uses $\zeta_{p_{k=0,1}} (S_{2n}^{i})$ as described in Equation~\ref{eqn:hyppc} and Equation~\ref{eqn:hypce}, respectively. The overall loss function used to train the pipeline is a combination of  the two losses and is defined as:
\begin{equation}\label{eqn:total_loss}
    \mathcal{L} = \mathcal{L}_{\text{Hyp-PC}} + \mathcal{L}_{\text{Hyp-CE}}.
\end{equation}
In our approach, we employ a Euclidean optimizer for training parameters in both Euclidean and hyperbolic spaces. Although the optimization process for hyperbolic parameters typically requires Riemannian gradients, when working in the conformal Poincaré Ball model, Riemannian gradients are equivalent to the Euclidean gradients with a scaling factor~\cite{nickel2017poincare}. This allows for the utilization of standard back-propagation techniques in our computations. To stabilize the training of the hyperbolic one-class classifier we perform feature clipping~\cite{guo2022clipped} in the Euclidean space that addresses the issue of vanishing gradients found in the hyperbolic space. Furthermore, due to the exponential nature of geodesic distances near the edges of the Poincar\'e Ball, we perform gradient clipping to avoid exploding gradients and regularize the parameter updates. The complete training framework is formalized in Algorithm~\ref{alg:hypoc}.

\input{figures/dataset}

%% file: figures/poincare.tex
\begin{figure}[t]
  \centering
   \includegraphics[width=0.9\linewidth]{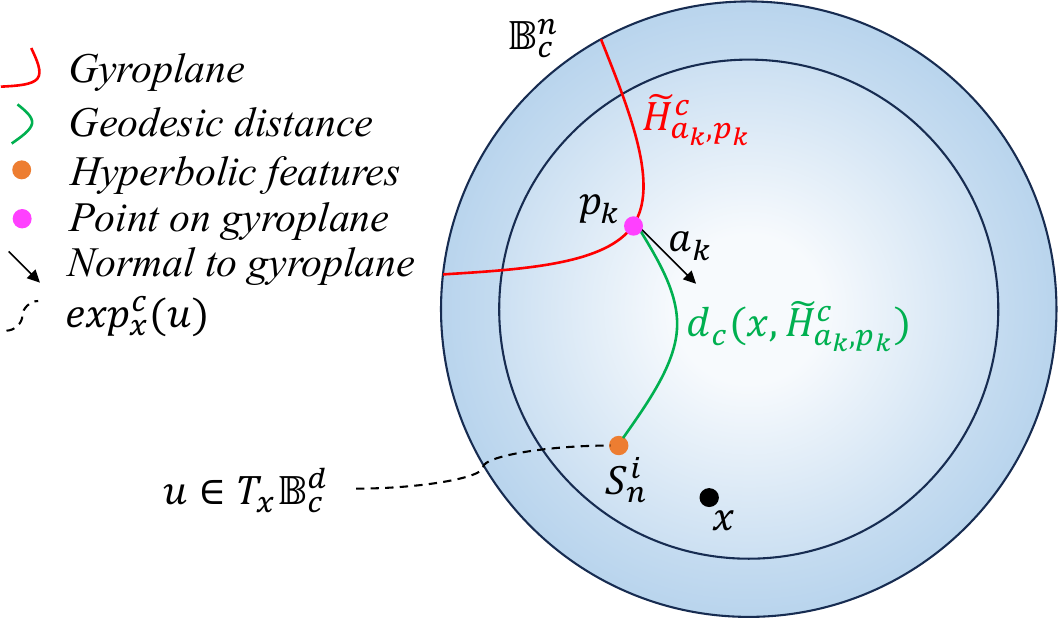}
   \caption{Visualization of the Poincar\`e Ball $\mathbb{B}_{c}^d$. $S_n^i$ denotes hyperbolic features exponentially mapped from $\mathcal{T}_x\mathbb{B}^d_c$. In our work, we use $d_c(x, \Tilde{H}_{a_k,p_k}^c)$ to compute $\mathcal{L}_{Hyp-PC}$ and $\mathcal{L}_{Hyp-CE}$. $\Tilde{H}_{a,p}^c$ represents the \textit{gyroplane} of class $k$.} 
   \label{fig:poincare}
   \vspace{-4pt}
\end{figure}

%% file: figures/framework.tex
\begin{figure*}[t]
  \centering
   \includegraphics[width=\linewidth]{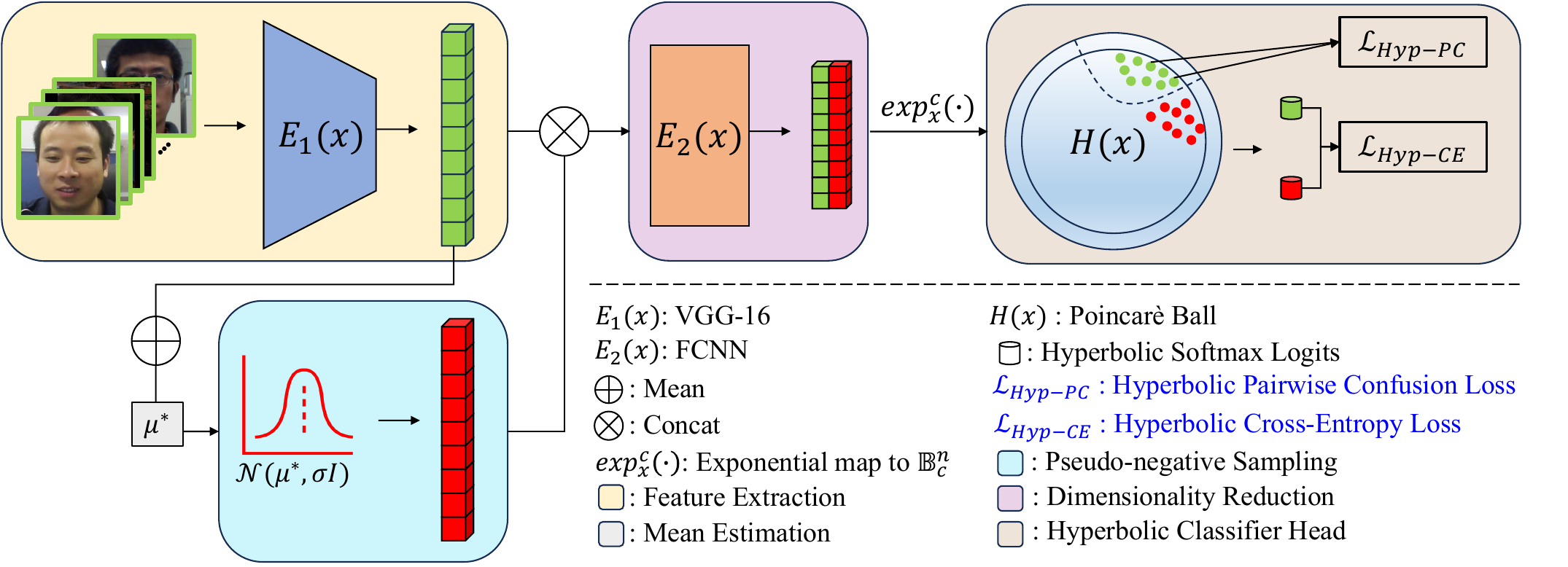}
     \caption{Overview of the proposed pipeline \textcolor{blue}{Hyp-OC} (Section~\ref{sec:training}). $E_1(x)$ extracts the facial features. The facial features are used to estimate the mean of Gaussian distribution utilized to sample pseudo-negative points. The real features and pseudo-negative features are then concatenated and passed to $E_2(x)$ for dimensionality reduction. The low-dimension features are mapped to Poincar\'e Ball using \textit{exponential map}. The training objective is to minimize the summation of the \textcolor{blue}{proposed} loss functions \textcolor{blue}{$\mathcal{L}_{Hyp-PC}$} (Section~\ref{sec:Hyp-PC}) and \textcolor{blue}{$\mathcal{L}_{Hyp-CE}$} (Section~\ref{sec:Hyp-CE}). The result is a separating \textit{gyroplane} beneficial for one-class face anti-spoofing. [Best viewed in color]}
   \label{fig:framework}
\end{figure*}

%% file: sec/algorithm.tex
\begin{algorithm}[!ht]
\begin{algorithmic}[1]
  \small
  \State {\textbf{Input:}} Training data $\mathcal{D}=\{(\mathbb{X}_n,\mathbb{Y}_n)_{i}\}_{i=1}^N$, facial feature extractor $\text{E}_1(x)$, FCNN  $\text{E}_2(x)$ for dimensionality reduction, Poincar\'e Ball $(\mathbb{B}_{c}^{d}, g^{\mathbb{B}_{c}})$ with manifold $\mathbb{B}_{c}^{d} = \{x \in \mathbb{R}^{d} : c\lVert x \rVert < 1, c \geq 0\}$, hyperbolic classifier $H(x)$ i.e. $\Tilde{H}_{a,p}^c := \{x \in \mathbb{B}_c^{d} \colon \langle -p \oplus_c x, a\rangle = 0 \}$ for $p \in \mathbb{B}_c^{d}$, $a \in \mathcal{T}_{p}\mathbb{B}_c^{d} \backslash \{0\}$, learning rate $\gamma$, Adam optimizer momentum parameters - $\beta_1 \& \beta_2$, hyper-parameter $\alpha$ for adaptive mean, Euclidean feature clipping function $\text{feat\_clip}(,r)$, Gradient clipping function $\text{grad\_clip}(,p)$, Gaussian distribution $\mathcal{N}(\mu, \sigma I); \mu,\sigma \in \mathbb{R}^{d^\prime}$ to sample pseudo-negative points, total epochs $T = N/n \ni 2 \mid n$.
  \State {\textbf{Init:}} $c=0.1$, $\alpha = 0.8$, $\beta_1 = 0.9$, $\beta_2 = 0.999$, $r=2$, $p=3$, $\mu=\textbf{0}$, $\sigma=\textbf{1}$, $d=4096$, $d = 128$ 
    \For{$\text{t in 0, 1, ..., T}$}
        \State $----------\text{E}_1(x)----------$
        \State Input Data: $\mathbb{X}_n^t = \{X_n^i \mid 0 \leq i < n\}$.
        \State $\mathbb{F}_n^t = \{F_n^i \mid 0 \leq i < n, \dim(F_n^i) = d^\prime\} = \text{E}_1(\mathbb{X}_n^t)$.
        \State \textcolor{blue}{\# Estimating mean of $\mathcal{N}(\mu, \sigma I)$}
        \State $\mu_{\text{current}} = \sum_{i=0}^{n} X_{n}^i$.
        \State $\mu^t = \alpha\mu^{t-1} + (1-\alpha)\mu_{\text{current}}$.
        \State $\mathbb{P}_n^t$: Sample $n$ pseudo-negative points from $\mathcal{N}(\mu, \sigma I)$.
        \State $\mathbb{X}_{2n}^t := \mathbb{X}_{n}^t \; concat \; \mathbb{P}_n^t$
        \State $----------\text{E}_2(x)----------$
        \State Input Data: $\mathbb{X}_{2n}^t = \{X_{2n}^i \mid 0 \leq i < 2n\}$.
        \State $\mathbb{G}_{2n}^t=\{G_{2n}^i \mid 0 \leq i < 2n, \dim(G_{2n}^i) = d \}$
        \State $\:\:\:\:\:\:\:\:\: = E_2(\mathbb{X}_{2n}^t)$.
        \State $\mathbb{G}_{2n}^t \gets \text{feat\_clip}(f, r) \; \forall f \in \mathbb{G}_{2n}^t$.
        \State \textcolor{blue}{\# Mapping  $\mathbb{G}_{2n}^t$ to $(\mathbb{B}_{c}^{d}, g^{\mathbb{B}_{c}})$}
        \State $\mathbb{G}_{2n}^t \xRightarrow[\text{\textit{exp map}}]{\exp_{x^t}^c} \mathbb{S}_{2n}^t; \; \exp_{x^t}^c : \mathcal{T}_{x^t}\mathbb{B}^{d}_c \to \mathbb{B}^{d}_c;\; x^t \in \mathbb{B}^{d}_c$.
        \State $----------\text{H}(x)----------$
        \State Input Data: $\mathbb{S}_{2n}^t = \{S_{2n}^{i} \mid 0\leq i<2n\}$.
        \State \textcolor{blue}{\# Calculating hyperbolic-softmax logits} 
        \State $\zeta_{p_k} (S_{2n}^{t}) = H(\mathbb{S}_{2n}^t)$; \;\; $k=0,1$.
        \State $-----------------------$
        \State \textbf{Loss calculation:} $\mathcal{L} = \mathcal{L}_{\text{Hyp-PC}} + \mathcal{L}_{\text{Hyp-CE}}$.
        \State \textcolor{blue}{\# Trainable parameters}
        \State $\mathcal{W}^t$ = \{$W_{\text{E}_1}^t$, $W_{\text{E}_2}^t$, $W_\text{H}^t$\}.
        \State \textbf{Model update:}
        \For{$w^t \in \mathcal{W}^t$}
            \If {$w^t \in \{W_{\text{E}_1}^t$, $W_{\text{E}_2}^t\}$}
            \State \textcolor{orange}{$ g^t = \text{grad\_clip}(\nabla_{w^t}\mathcal{L},r)$}
            \ElsIf {$w^t \in \{W_{\text{H}}^t \}$}
            \State \textcolor{red}{$ g^t = \text{grad\_clip}(\nabla_{w^t}\mathcal{L},r) \cdot \frac{(1 - c\lVert x^t \rVert)^2}{4}$}
            \EndIf
            \State $m^t = \beta_1 \cdot m^{t-1} + (1 - \beta_1) \cdot g^t$
            \State $v^t = \beta_2 \cdot v^{t-1} + (1 - \beta_2) \cdot {g^t}^2$
            \State $\hat{m}^t = m^t / (1 - \beta_1^t)$
            \State $\hat{v}^t = v^t / (1 - \beta_2^t)$
            \State ${w}^{t+1} = w^{t} - \gamma \cdot \frac{\hat{m}^t}{(\sqrt{\hat{v}^t} + \epsilon)}$
        \EndFor
    \EndFor
\end{algorithmic} 
\small{
    \textcolor{blue}{blue}: comments\\
    \textcolor{orange}{orange}: euclidean parameter updates\\
    \textcolor{red}{red}: hyperbolic parameter updates}
\caption{Training Framework}
\label{alg:hypoc}
\end{algorithm}

%% file: figures/dataset.tex
\begin{figure*}[t]
  \centering
   \includegraphics[width=\linewidth]{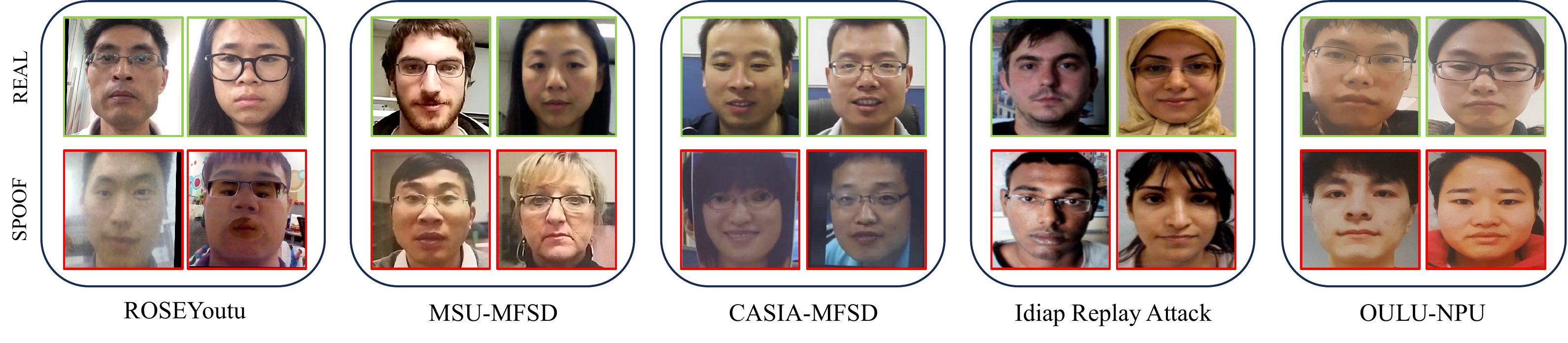}
    \caption{Sample images from the datasets used for training: RoseYoutu~\cite{li2022one}, MSU-MFSD~\cite{wen2015face}, CASIA-MFSD~\cite{zhang2012face}, Idiap Replay Attack~\cite{chingovska2012effectiveness}, OULU-NPU~\cite{boulkenafet2017oulu}.}
   \label{fig:dataset}
   \vspace{-8pt}
\end{figure*}

%% file: sec/4_experiments.tex
\section{Experiments}
This section, describes the datasets and protocols used to evaluate our proposed pipeline. We give a brief account of the evaluation metrics and the baseline methods for comparison. Finally, we detail the implementation steps taken to train our proposed model.
\subsection{Datasets and Protocols}
We evaluate our proposed hyperbolic one-class classification framework using three different protocols. The protocols holistically assess the model for intra-domain and inter-domain performance and demonstrate the superiority of the proposed approach. In \textbf{Protocol 1}, we evaluate intra-domain performance where the network is trained and tested on the same dataset. We employ widely used FAS datasets: ROSE-Youtu \textbf{R}~\cite{li2022one}, MSU-MFSD \textbf{(M)}~\cite{wen2015face}, CASIA-MFSD \textbf{(C)}~\cite{zhang2012face}, Idiap Replay Attack \textbf{(I)}~\cite{chingovska2012effectiveness}, and OULU-NPU \textbf{(O)}~\cite{boulkenafet2017oulu}. Some sample images of the employed datasets are shown in Figure~\ref{fig:dataset}. In \textbf{Protocol 2}, we use the \textbf{MCIO} datasets and follow the leave-one-out setting. In particular, a model is trained on multiple source domains and tested on a single target domain. For instance, \textbf{MCI} $\to$ \textbf{O} represents that the model is trained on \textbf{M},\textbf{C} and \textbf{I} and tested on \textbf{O}. In \textbf{Protocol 3}, we again use \textbf{M}, \textbf{C}, \textbf{I}, and \textbf{O} datasets and follow a single-source-single-target setting. In particular, a model is trained using a single source domain and tested on a single target domain different from the source domain. Protocols 2 and 3 evaluate the inter-domain performance of the model in two different settings. We only focus on single-modal FAS datasets for all experiments, as discussed in Section~\ref{sec:intro}.

\subsection{Performance Metrics and Baseline Methods}
We evaluate the model performance using the standard metrics for FAS: Attack Presentation Classification Error Rate (APCER), Bonafide Presentation Classification Error Rate (BPCER), Half Total Error Rate (HTER), and Area under the ROC curve (AUC). We run experiments for Protocol-1 and Protocol-3 five times and Protocol-2 three times and report the mean. We compare our model with seven state-of-the-art one-class classifiers: OC-SVM~\cite{scholkopf2001estimating}, OC-GMM~\cite{ilonen2006gaussian}, SVDD~\cite{tax2004support}, MD~\cite{nader2014mahalanobis}, OC-CNN~\cite{oza2018one}, Anomaly Detection-based unknown fPAD~\cite{baweja2020anomaly} and DROCC~\cite{goyal2020drocc}. These classifiers have been previously used for FAS and are the closest to our proposed work. We re-train all the baselines to make the dataset protocol consistent with our work. We employ the same feature extractor as Hyp-OC for all the baseline models.

\subsection{Implementation Details} 
The input to all the models are RGB images, resized and center cropped to $224 \times 224 \times 3$. We normalize the images with mean $[129.186, 104.762, 93.593]$ and standard deviation $[1, 1, 1]$. We use VGG-16 pre-trained on VGGFace~\cite{parkhi2015deep} as our feature extractor for all models. We employ a fully connected neural network for dimensionality reduction. The FCNN consists of 3 hidden layers with $8192$, $1000$, and $512$ neurons, respectively. The input and output layers have dimensions $4096$ and $128$, respectively. We perform the \textit{exponential map} of Euclidean features to Poincar\'e Ball model of the Hyperbolic space of dimension $128$. For all the baselines and proposed method, we perform gradient updates on all FCNN and Poincar\'e Ball parameters. We only update the weights and biases of the last $3$ convolution layers and the last fully connected layer of the VGG-16 feature extractor for all the models except DROCC. For DROCC~\cite{goyal2020drocc}, we update all the parameters of the VGG-16 feature extractor, which gives better results than only updating the last layers but still falls short compared to the proposed Hyp-OC. 

For training of the models, we use Adam optimizer with a learning rate of 1\text{e}-6, momentum parameters as $(0.9,0.999)$, and weight decay set to 1\text{e}-6. We use different batch sizes and training epochs for each dataset. For \textbf{R}, \textbf{M}, \textbf{C}, \textbf{I}, \textbf{O}, we use a batch size of $8,8,8,32,32$ and train it for $60,100,100,50,60$ epochs, respectively. We train the models on 8 NVIDIA A5000 GPUs, each with 24GB memory. The training steps of our proposed approach, as shown in Algorithm~\ref{alg:hypoc}, highlights that gradients of hyperbolic parameters are scaled versions of Euclidean parameters~\cite{nickel2017poincare} and are dependent on the curvature of the hyperbolic space.

We re-train all the baselines to make the dataset protocol consistent for a fair comparison. We use the OneClassSVM implementation from the sklearn library~\cite{pedregosa2011scikit} with `rbf' kernel and $\nu=0.1$. We use the signed distance from the hyperplane to calculate the performance metrics. We follow the standard sklearn implementation of OC-GMM~\cite{ilonen2006gaussian} with $\text{n\_components}=1$. We take the log-likelihood of a sample belonging to GMM as the score to calculate performance metrics. We use SVDD implementation of~\cite{svdd}, and chose the kernel `rbf', $C=0.9$, and $\gamma=0.3$. The distance from the center is used to calculate the metrics. For implementing MD~\cite{nader2014mahalanobis}, we again follow the standard implementation of the sklearn library, using the distance itself to calculate metrics. For OC-CNN~\cite{oza2018one}, AD-fPAD~\cite{baweja2020anomaly}, and DROCC~\cite{goyal2020drocc}, we follow the official GitHub implementation and use the probability of a sample belonging to the spoof class as the score to calculate performance metrics.

%% file: sec/5_results.tex
\section{Results and Analysis}
\input{tables/protocol1}
In this section, we compare our proposed approach with recent works that utilize one-class classifiers for FAS. Table~\ref{tab:protocol1} reports the intra-domain testing performance for \textbf{Protocol 1}. Table~\ref{tab:protocol2} and Table~\ref{tab:protocol3} outline the inter-domain performance for \textbf{Protocol 2} and \textbf{Protocol 3}, in leave-one-out setting and single-source-single-target setting, respectively. Our proposed approach significantly outperforms previous methods. Specifically, we improve upon the best baseline by an Avg. HTER of $7.493$ in Protocol-1, $4.231$ in Protocol-2, and $2.778$ in Protocol-3.

\input{tables/protocol3}

\subsection{Protocol 1}
\label{sec:protocol1}
Table~\ref{tab:protocol1} presents the results of Protocol 1, illustrating that Hyp-OC significantly outperforms previous baselines in four out of five datasets. In particular, it shows enhancements of $7.733$ on \textbf{R}, $3.333$ on \textbf{M}, $14.518$ on \textbf{C}, and $9.506$ on \textbf{I} in terms of HTER. However, it is the second-best performer on \textbf{O}, $4.777$ behind the top baseline. Overall, Hyp-OC demonstrates superior performance across all datasets with an Avg. HTER of $28.339$, marking a $7.493$ increase upon the best baseline. The AUC scores of Hyp-OC are $0.713$, $0.782$, $0.784$, $0.931$, and $0.639$ on \textbf{R}, \textbf{M}, \textbf{C}, \textbf{I}, and \textbf{O}, respectively. The intra-domain testing results establish Hyp-OC as the new state-of-the-art one-class face anti-spoofing model.

\subsection{Protocol 2}
\input{tables/protocol2}
\label{sec:protocol2}
Protocol 2 highlights the capability of Hyp-OC to generalize over unseen environments. From Table~\ref{tab:protocol2}, it can be inferred that Hyp-OC performs better than other baselines by a huge margin. Hyp-OC exhibits remarkable proficiency in modeling the real class of different domains irrespective of the diverse changes in environmental factors such as lighting, camera angles, and backgrounds of each domain. When compared to previous baselines in terms of HTER, Hyp-OC achieves better results in three out of the four evaluated target domains. The performance improvement on \textbf{OCI} $\to$ \textbf{M}, \textbf{OMI} $\to$ \textbf{C} and \textbf{OCM} $\to$ \textbf{I} are $0.417$, $7.747$ and $3.872$ respectively. Hyp-OC achieves an overall performance gain of $4.231$ upon the best baseline.

\subsection{Protocol 3}
\label{sec:protocol3}
Table~\ref{tab:protocol3} summarizes the findings for Protocol 3, an inter-domain protocol that assesses a model's performance across multiple domains while being trained on a single domain. This protocol serves to highlight the generalizability of the learned features. Hyp-OC outperforms previous baselines in six out of twelve single-domain-single-target experiments. For target domain \textbf{C}, a huge improvement in HTER is observed, with gains of $12.555$, $9.000$, and $8.000$, respectively. We observe that SVDD~\cite{tax2004support} performs better than Hyp-OC on the target domain \textbf{M}. Also, Hyp-OC doesn't perform as good on target domain \textbf{O}, and is exceeded by AD-fPAD~\cite{baweja2020anomaly} on \textbf{C} $\to$ \textbf{O}, OC-CNN~\cite{oza2018one} on \textbf{I} $\to$ \textbf{O} and SVDD~\cite{tax2004support} on \textbf{M} $\to$ \textbf{O}. The overall Avg. HTER of Hyp-OC across all experiments is $38.196$, which surpasses all baselines.    

\subsection{Ablation and Analysis}

In the ablation study, we analyze the influence of each component of the proposed pipeline on the performance. Furthermore, we discuss the effect of different values for the Euclidean feature clipping and different curvature values of the hyperbolic space on the FAS performance. In the end, we analyze the use of Hyp-OC with other feature extractors. 

\noindent\textbf{Proposed One-class classifier for FAS:}
The effect of different components in the proposed pipeline is shown in Table~\ref{tab:ablation}. There are two major components that stand out and impact the performance - Adaptive mean (Table~\ref{tab:ablation} row-1) and Hyp-OC (Table~\ref{tab:ablation} row-2). The absence of \textbf{adaptive mean} from the pipeline results in a  drop of 14.885 in HTER. Adaptive mean helps to estimate the mean of the Gaussian distribution used to sample pseudo-negative points. In FAS, the spoof samples lie close to the real samples in the feature space. The adaptive mean strategy pushes the mean of Gaussian distribution towards the cluster of real samples, thus helping to accurately sample pseudo-negative points. This helps improve the overall performance of the pipeline. 
The absence of \textbf{Hyp-OC} results in a noticeable decline in performance, with HTER dropping by 7.493. This shows that the Poincar\`e Ball model effectively embeds feature representations for FAS task. Hyp-OC allows for better fitting of separating \textit{gyroplane} for one-class classification. 
The geodesic distance increases exponentially near the boundary of the Poincar\`e Ball. \textbf{Euclidean feature clipping} helps to cut down the effective radius of the Poincar\'e Ball and solves the problem of vanishing gradients persistent in Hyperbolic space. The clipping stabilizes the training and acts as a regularizer. This is validated by the performance drop (Table~\ref{tab:ablation} row-3) of 3.707 in HTER when Euclidean feature clipping is removed from the pipeline. Lastly, the proposed loss function \textbf{Hyp-PC} disrupts the feature space by creating confusion. It removes the identity information by unclustering the feature representations, making it better suited for FAS task. Additionally, it improves the mean estimation for the pseudo-negative Gaussian distribution. The absence of Hyp-PC loss results in a performance drop (Table~\ref{tab:ablation} row-4) of $2.569$ in HTER, signifying its importance in the pipeline. 

\input{tables/ablation}
\input{figures/ablation}

\noindent \textbf{Curvature of the Hyperbolic space:}
We perform experiments for different curvatures of the Poincar\'e Ball and compare the Avg. HTER as shown in Figure~\ref{fig:ablation} (left). We can see that the performance is better at lower curvature values. Moreover, we observe that the difference between APCER and BPCER is less at low curvature values than when the curvature is high. It indicates that the training is more stable at lower curvatures of hyperbolic spaces. In our implementation, following previous works~\cite{ermolov2022hyperbolic, atigh2022hyperbolic}, we choose the curvature of the hyperbolic space to be 0.1. 
\noindent \textbf{Euclidean feature clipping value:}
We implemented Euclidean feature clipping~\cite{guo2022clipped} to regularize our training in the hyperbolic space. The clipping value directly relates to the effective radius of the Poincar\'e Ball. We perform experiments for different clipping values and observe that the performance initially increases and then decreases slowly, as shown in Figure~\ref{fig:ablation} (right). This is in agreement with ~\cite{guo2022clipped}, which shows that the effective radius increases from $0 \to \approx0.8$ when we increase the clipping value from $0 \to 1$ and $\lim_{\text{clipping value} \to \infty} \text{effective radius} = 1$. In our work, we choose the clipping radius to be $2$, which gives the best performance.

\noindent \textbf{Feature extractor:}
To support our choice of VGGFace~\cite{parkhi2015deep} as a feature extractor, we perform experiments with different feature extractors and compare the Avg. HTER. We use ResNet50~\cite{he2016deep} , SeNet50~\cite{hu2018squeeze} and CLIP ViT~\cite{radford2021learning} for comparison and achieve Avg. HTER of $33.555$, $37.915$ and $39.505$, respectively. We observe that VGGFace performs better in three out of five datasets. However, ResNet50 performs better on \textbf{R}, and CLIP ViT performs better on \textbf{O}. The results indicate that VGGFace is a better choice of feature extractor for one-class FAS using hyperbolic embeddings.
\input{tables/ablation_feature_extractor}

%% file: tables/protocol1.tex
\begin{table*}[t]
  \centering
  \scalebox{1}{
  \begin{tabular}{@{} lcccccccccc>{\columncolor[gray]{0.9}}c}
    \toprule
        \multirow{2}{*}{\textbf{Method}} & \multicolumn{2}{c}{\textbf{ROSEYoutu}} & \multicolumn{2}{c}{\textbf{MSU-MFSD}} & \multicolumn{2}{c}{\textbf{CASIA-MFSD}} & \multicolumn{2}{c}{\textbf{ReplayAttack}} &  \multicolumn{2}{c}{\textbf{OULU-NPU}} & \multicolumn{1}{c}{\textbf{Avg.}}\\
        \cmidrule{2-3}  \cmidrule{4-5} \cmidrule{6-7} \cmidrule{8-9} \cmidrule{10-11} \cmidrule{12-12}
        \multicolumn{1}{c}{} & \multicolumn{1}{c}{\textbf{HTER }$\downarrow$} & \multicolumn{1}{c}{\textbf{AUC} $\uparrow$} & \multicolumn{1}{c}{\textbf{HTER }$\downarrow$} & \multicolumn{1}{c}{\textbf{AUC} $\uparrow$} & \multicolumn{1}{c}{\textbf{HTER }$\downarrow$} & \multicolumn{1}{c}{\textbf{AUC} $\uparrow$} & \multicolumn{1}{c}{\textbf{HTER }$\downarrow$} & \multicolumn{1}{c}{\textbf{AUC} $\uparrow$} & \multicolumn{1}{c}{\textbf{HTER }$\downarrow$} & \multicolumn{1}{c}{\textbf{AUC} $\uparrow$} & \multicolumn{1}{c}{\textbf{HTER }$\downarrow$}\\
    \midrule
    OC-SVM~\cite{scholkopf2001estimating} &  56.843 & 0.415 & 54.375 & 0.509 & 47.778 & 0.582 & 59.892 & 0.401 & 57.323 & 0.426 & 55.242\\
    OC-GMM~\cite{ilonen2006gaussian} &  62.427 & 0.302 & 50.000 & 0.268 & 62.593 & 0.324 & 60.158 & 0.054 & 54.948 & 0.458 & 58.025\\
    SVDD~\cite{tax2004support} & 41.807 & 0.625 & 36.250 & 0.705 & 40.648 & 0.653 & 24.633 & 0.838 & \textcolor{blue}{\textbf{35.823}} & 0.682 & 35.832\\
    MD~\cite{nader2014mahalanobis} & 51.413 & 0.474 & 49.583 & 0.512 & 57.963 & 0.362 & 47.137 & 0.576 & 58.010 & 0.408 & 52.822\\
    OC-CNN~\cite{oza2018one} & 46.865 & 0.538 & 37.292 & 0.674 & 44.722 & 0.584 & 34.825 & 0.716 & 44.302 & 0.561 & 41.601\\
    AD-fPAD~\cite{baweja2020anomaly} & 43.157 & 0.569 & 30.625 & 0.733 & 39.537 & 0.651 & 24.217 & 0.824 & 41.625 & 0.605 & 35.832\\
    DROCC~\cite{goyal2020drocc} & 52.865 & 0.444 & 48.542 & 0.499 & 48.333 & 0.499 & 40.592 & 0.648 & 41.906 & 0.608 & 46.448 \\
    \midrule
    Ours & \textcolor{blue}{\textbf{34.074}} & 0.713 & \textcolor{blue}{\textbf{27.292}} & 0.782 & \textcolor{blue}{\textbf{25.019}} & 0.784 & \textcolor{blue}{\textbf{14.711}} & 0.931 & 40.600 & 0.639 & \textcolor{blue}{\textbf{28.339}}\\
    \bottomrule
  \end{tabular}}
  \caption{Results of intra-domain performance in \textbf{Protocol 1}. We run each experiment five times and report the mean HTER and AUC.  }
  \label{tab:protocol1}
  \vspace{-4pt}
\end{table*}

%% file: tables/protocol3.tex
\begin{table*}[t]
  \centering
  \scalebox{0.93}{
  \begin{tabular}{@{} lcccccccccccc>{\columncolor[gray]{0.9}}c}
    \toprule
        \textbf{Method} & 
        \textbf{C} $\rightarrow$ \textbf{I} & 
        \textbf{C} $\rightarrow$ \textbf{M} & 
        \textbf{C} $\rightarrow$ \textbf{O} &
        \textbf{I} $\rightarrow$ \textbf{C} & 
        \textbf{I} $\rightarrow$ \textbf{M} & 
        \textbf{I} $\rightarrow$ \textbf{O} &
        \textbf{M} $\rightarrow$ \textbf{C} & 
        \textbf{M} $\rightarrow$ \textbf{I} & 
        \textbf{M} $\rightarrow$ \textbf{O} &
        \textbf{O} $\rightarrow$ \textbf{C} & 
        \textbf{O} $\rightarrow$ \textbf{I} & 
        \textbf{O} $\rightarrow$ \textbf{M} &
        \multicolumn{1}{c}{\textbf{Avg.}}\\
    \midrule
    OC-SVM~\cite{scholkopf2001estimating} & 36.033 & 34.167 & 41.583 & 51.667 & 37.500 & 43.917 & 45.648 & 45.696 & 55.865 & 55.093 & 53.354 & 45.000 & 45.460\\
    OC-GMM~\cite{ilonen2006gaussian} & 50.000 & 50.625 & 50.000 & 53.426 & 61.667 & 51.792 & 50.000 & 50.000 & 50.000 & 50.463 & 50.888 & 65.000 & 52.822\\
    SVDD~\cite{tax2004support} & 35.225 & \textcolor{blue}{\textbf{33.958}} & 40.729 & 41.944 & 37.500 & 45.031 & 44.630 & 50.013 & \textcolor{blue}{\textbf{37.469}} & 45.926 & 46.967 & \textcolor{blue}{\textbf{32.292}} & 40.974\\
    MD~\cite{nader2014mahalanobis} & 47.246 & 56.042 & 58.427 & 58.148 & 54.792 & 58.552 & 58.426 & 47.175 & 56.781 & 58.611 & 46.929 & 54.167 & 54.608\\
    OC-CNN~\cite{oza2018one} & 49.363 & 49.375 & 39.344 & 48.519 & 48.750 & \textcolor{blue}{\textbf{42.552}} & 49.722 & 56.067 & 49.469 & 47.130 & 44.767 & 45.833 & 47.574\\
    AD-fPAD~\cite{baweja2020anomaly} & 50.192 & 47.917 & \textcolor{blue}{\textbf{35.208}} & 40.278 & 44.792 & 45.667 & 48.796 & 48.408 & 48.531 & 50.463 & 54.017 & 45.625 & 46.658\\
    DROCC~\cite{goyal2020drocc} & 47.542 & 70.000 & 52.958 & 41.667 & 53.750 & 46.177 & 47.778 & 50.483 & 40.875 & 45.000 & \textcolor{blue}{\textbf{35.129}} & 47.292 & 48.221\\
    \midrule
    Ours & \textcolor{blue}{\textbf{29.672}} & 35.667 & 45.217 & \textcolor{blue}{\textbf{32.278}} & \textcolor{blue}{\textbf{36.833}} & 46.254 & \textcolor{blue}{\textbf{33.093}} & \textcolor{blue}{\textbf{35.790}} & 41.229 & \textcolor{blue}{\textbf{36.000}} & 51.360 & 34.958 & \textcolor{blue}{\textbf{38.196}}\\
    \bottomrule
  \end{tabular}}
  \caption{Results of inter-domain performance (single-source-single-target setting) in \textbf{Protocol 3}. The domains used are MSU-MFSD (\textbf{M}), CASIA-MFSD (\textbf{C}), Idiap Replay Attack (\textbf{I}) and OULU-NPU (\textbf{O}). We run each experiment five times and report the mean HTER.}
  \label{tab:protocol3}
  \vspace{-8pt}
\end{table*}

%% file: tables/protocol2.tex
\begin{table}
  \centering
  \scalebox{0.85}{
  \begin{tabular}{@{} lcccc>{\columncolor[gray]{0.9}}c}
    \toprule
        \multirow{1}{*}{\textbf{Method}} & \multicolumn{1}{c}{\textbf{OCI} $\rightarrow$ \textbf{M}} & \multicolumn{1}{c}{\textbf{OMI} $\rightarrow$ \textbf{C}} & \multicolumn{1}{c}{\textbf{OCM} $\rightarrow$ \textbf{I}} & \multicolumn{1}{c}{\textbf{ICM} $\rightarrow$ \textbf{O}} & \multicolumn{1}{c}{\textbf{Avg.}}
    \\
    \midrule
    OC-SVM~\cite{scholkopf2001estimating} & 43.333 & 54.352 & 52.433 & 42.167 & 48.071\\
    OC-GMM~\cite{ilonen2006gaussian} & 67.083 & 50.185 & 50.250 & 56.083 & 55.900\\
    SVDD~\cite{tax2004support} & 32.292 & 43.981 & 48.250 & \textcolor{blue}{\textbf{36.125}} & 40.162\\
    MD~\cite{nader2014mahalanobis} & 54.167 & 58.611 & 47.058 & 58.500 & 54.584\\
    OC-CNN~\cite{oza2018one} & 44.375 & 49.722 & 38.471 & 48.365 & 45.233\\
    AD-fPAD~\cite{baweja2020anomaly} & 36.250 & 38.025 & 34.650 & 44.087 & 38.253\\
    DROCC~\cite{goyal2020drocc} & 37.917 & 42.037 & 43.250 & 45.656 & 42.215\\
    \midrule
    Ours & \textcolor{blue}{\textbf{31.875}} & \textcolor{blue}{\textbf{30.278}} & \textcolor{blue}{\textbf{30.778}} & 43.156 & \textcolor{blue}{\textbf{34.022}}\\
    \bottomrule
  \end{tabular}}
  \caption{Results of inter-domain performance (leave-one-out setting) in \textbf{Protocol 2}. The domains used are MSU-MFSD (\textbf{M}), CASIA-MFSD (\textbf{C}), Idiap Replay Attack (\textbf{I}) and OULU-NPU (\textbf{O}). We run each experiment thrice and report the mean HTER.}
  \label{tab:protocol2}
  \vspace{-14pt}
\end{table}

%% file: tables/ablation.tex
\begin{table}
  \centering
  \scalebox{1}{
  \begin{tabular}{cccc>{\columncolor[gray]{0.9}}c}
    \toprule
        \multicolumn{1}{c}{\begin{tabular}[c]{@{}c@{}}\textbf{Adaptive}\\ \textbf{mean}\end{tabular}} & \multicolumn{1}{c}{\begin{tabular}[c]{@{}c@{}}\textbf{Euclidean} \\ \textbf{feature clipping}\end{tabular}} & \multicolumn{1}{c}{\textbf{Hyp-OC}} & \multicolumn{1}{c}{\textbf{Hyp-PC}} & \multicolumn{1}{c}{\begin{tabular}[c]{@{}c@{}}\textbf{Avg.}\\ \textbf{HTER}\end{tabular}}
    \\
    \midrule
    \xmark & \cmark & \cmark & \cmark & 43.224\\
    \cmark & \xmark & \xmark & \xmark & 35.832\\
    \cmark & \xmark & \cmark & \cmark & 32.046\\
    \cmark & \cmark & \cmark & \xmark & 30.908\\
    \cmark & \cmark & \cmark & \cmark & \textcolor{blue}{\textbf{28.339}}\\
    \bottomrule
  \end{tabular}}
  \caption{Impact of various components of the proposed pipeline on performance. We report Avg. HTER for \textbf{Protocol 1}.}
  \label{tab:ablation}
\end{table}

%% file: figures/ablation.tex
\begin{figure}[t]
  \centering
   \includegraphics[width=\linewidth]{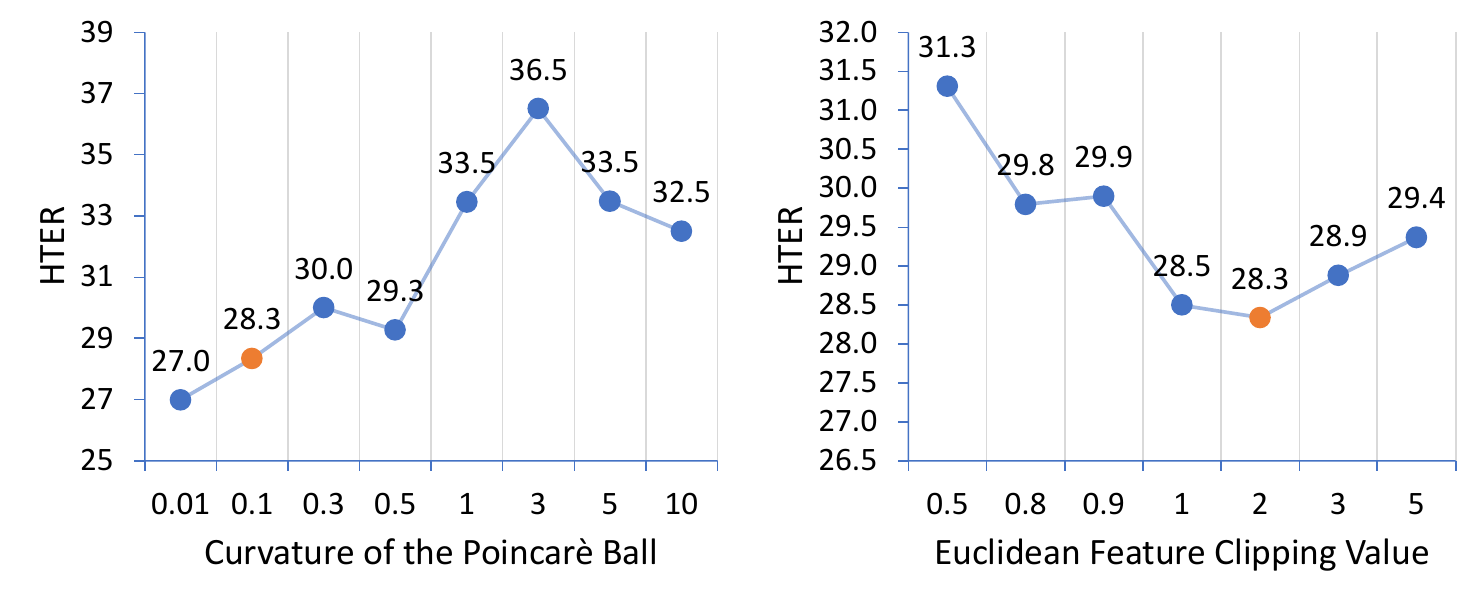}
   \caption{(Left) HTER performance w.r.t different curvatures of the Poincar\'e Ball. In our work, we fix the curvature of Poincar\'e Ball to $0.1$ (orange). (Right) HTER performance w.r.t different Euclidean feature clipping values. In our work, we set the Euclidean feature clipping value to 2 (orange). }
    \label{fig:ablation}
    \vspace{-8pt}
\end{figure}

%% file: tables/ablation_feature_extractor.tex
\begin{table}[t]
  \centering
  \scalebox{0.93}{
  \begin{tabular}{@{} lccccc>{\columncolor[gray]{0.9}}c}
    \toprule
        \textbf{Method} & 
        \textbf{R}& 
        \textbf{M}& 
        \textbf{C}&
        \textbf{I} & 
        \textbf{O}& 
        \multicolumn{1}{c}{\textbf{Avg.}}\\
    \midrule
    ResNet50~\cite{he2016deep} & \textcolor{blue}{\textbf{27.898}} & 31.250 & 32.222 & 24.313 & 52.094 & 33.555\\
    SeNet50~\cite{hu2018squeeze} & 38.049 & 35.833 & 36.019 & 35.904 & 43.771 & 37.915\\
    CLIP ViT~\cite{radford2021learning} & 40.756 & 36.042 & 32.778 & 51.979 & \textcolor{blue}{\textbf{35.969}} & 39.505\\
    \midrule
    VGGFace~\cite{parkhi2015deep} & 34.074 & \textcolor{blue}{\textbf{27.292}} & \textcolor{blue}{\textbf{25.019}} & \textcolor{blue}{\textbf{14.711}} & 40.600 & \textcolor{blue}{\textbf{28.339}}\\
    \bottomrule
  \end{tabular}}
  \caption{Comparison of Avg. HTER performance using different feature extractors. We report the results for \textbf{Protocol 1}.}
  \label{tab:feature_extractor}
  \vspace{-12pt}
\end{table}

%% file: sec/6_conclusion.tex
\section{Conclusion and Future Work}
In this research, we redefine FAS as a one-class classification task. We discuss our motivation and showcase the significance of - the ``Why One-Class?" approach, emphasizing its practicality in real-world applications. We show the benefits of employing a hyperbolic classifier head (Hyp-OC) to develop a one-class classifier and demonstrate its effectiveness for FAS using three protocols. For training, we introduce two novel loss functions, Hyp-PC and Hyp-CE, that operate in the hyperbolic space. Our proposed pipeline outperforms previous baselines and sets a new benchmark for one-class FAS.

Despite our advancements over previous one-class FAS baselines, we recognize that the performance of Hyp-OC still lags behind that of binary classifiers. However, in real-world deployment, the distribution of spoof samples is considerably more complex than that of real samples. This complexity arises from the infinite variability in presentation attack instruments, which motivates our pursuit of OC-FAS. We advocate for the development of one-class classifiers using only real samples as a step towards creating truly generalized models that can detect a wide variety of spoof attacks. In the future, we plan to explore other ways to leverage hyperbolic embeddings to enhance FAS performance.

\textbf{Acknowledgement:} This work was supported by NSF CAREER award 2045489.